\documentclass[conference,a4paper]{IEEEtran}
\IEEEoverridecommandlockouts
\usepackage{cite}
\usepackage{amsmath,amssymb,amsfonts}
\usepackage{algorithmic}
\usepackage{graphicx}
\usepackage{textcomp}
\usepackage{xcolor}

\def\BibTeX{{\rm B\kern-.05em{\sc i\kern-.025em b}\kern-.08em
    T\kern-.1667em\lower.7ex\hbox{E}\kern-.125emX}}
\begin{document}

\title{Increasing the Diversity in RGB-to-Thermal Image Translation for Automotive Applications
\thanks{This research received funding from the Flemish Government (AI Research Program)
\scriptsize © 2024 IEEE. Personal use of this material is permitted.
    Permission from IEEE must be obtained for all other uses,
    in any current or future media, including reprinting/republishing
    this material for advertising or promotional purposes,
    creating new collective works, for resale or redistribution to servers or lists,
    or reuse of any copyrighted component of this work in other works.\\
    DOI: \texttt{10.1109/SENSORS60989.2024.10785056}

}}

\author{\IEEEauthorblockN{Kaili Wang$^{1*}$, Leonardo Ravaglia$^{1}$, Roberto Longo$^{1}$, Lore Goetschalckx$^{1}$, David Van Hamme$^{2}$, Julie Moeyersoms$^{1}$, \\
Ben Stoffelen$^{1}$, Tom De Schepper$^{1}$\\
$^{1}$ imec, 3001 Leuven, Belgium,
$^{2}$ imec-IPI-Ghent University, 9000 Ghent, Belgium\\
$^{*}$Corresponding author: kaili.wang@imec.be}
}

\maketitle

\begin{abstract}
Thermal imaging in Advanced Driver Assistance Systems (ADAS) improves road safety with superior perception in low-light and harsh weather conditions compared to traditional RGB cameras. However, research in this area faces challenges due to limited dataset availability and poor representation in driving simulators. RGB-to-thermal image translation offers a potential solution, but existing methods focus on one-to-one mappings. We propose a one-to-many mapping using a multi-modal translation framework enhanced with our Component-aware Adaptive Instance Normalization (CoAdaIN). Unlike the original AdaIN, which applies styles globally, CoAdaIN adapts styles to different image components individually. 
The result, as we show, is more realistic and diverse thermal image translations.
\end{abstract}


\section{Introduction}
Adaptive Driver Assistance Systems (ADAS) reduce human error behind the wheel by alerting drivers to obstacles, braking automatically, and more. To accurately represent its environment, an ADAS relies on inputs from different sensors, with traditional RGB cameras being the most common. In addition, thermal sensors are often adopted to improve vision in low-light conditions. Both for the development and evaluation of these systems, there is a need for sensor data covering diverse traffic scenarios. While RGB data is abundant, matching thermal data is limited. 

One approach to bridge this gap is to synthesize matching thermal images from their more readily available RGB counterparts using image-to-image (I2I) translation techniques~\cite{wang2018pix2pixHD}\cite{CycleGAN2017}. Prior efforts in this direction have relied on deterministic translation models, performing a one-to-one mapping~\cite{Kniaz_2018_ECCV_Workshops} \cite{Liu_2018_CVPR_Workshops}\cite{Iwashita_2019_CVPR_Workshops}\cite{Liu_HGGAN_2022}. However, one RGB image can correspond to multiple plausible thermal images. Imagine two identical hypothetical scenes featuring a parked car, except in one scene the car has only just been parked there. The fact that the car is still warm will yield a different thermal image for that scene, but not a different RGB image. Therefore, to generate truly rich and diverse thermal data, an I2I translation model should be able to synthesize many outputs with different ``styles" given a single input. To achieve this goal, the current work turns to a multi-modal I2I framework~\cite{huang2018munit}.

In this framework, different styles are applied through Adaptive Instance Normalization~(AdaIN)~\cite{huang2017adain}. However, the traditional AdaIN imposes a global style on the whole image, without differentiating between its components (e.g., scene objects). To synthesize even more diverse outputs and cover a broad spectrum of real-world scenarios, we propose a Component-aware Adaptive Instance Normalization (CoAdaIN). This improvement allows the translation model to apply different local styles (e.g., the temperature of one parked car), thereby extending the range of outputs it can produce.

In summary, the contribution is three-fold: 
1) We highlight one-to-one mapping as a shortcoming of current RGB-to-thermal translation methods and advocate for a one-to-many mapping approach instead. 
2) We propose Component-aware Adaptive Instance Normalization (CoAdaIN) to achieve component-level style variations in thermal images. 
3) We conduct both qualitative and quantitative evaluations of our generated images and showcase their diversity.

\section{Related Work}
\subsection{RGB2Thr translation}
A recent approach to RGB-to-thermal (RGB2Thr) translation entails a heatmap-guided model~\cite{Liu_HGGAN_2022} based on CycleGAN~\cite{CycleGAN2017}. Heatmaps representing key objects as well as Sobel edge maps are extracted to help the decoder reconstruct vital object information (e.g., sharp object edges). 
Similarly, variants of CycleGAN have been adopted in \cite{escyclegan_2019_sun}. 
The strength of a CycleGAN-based approach is that it alleviates the need for training pairs that are matched between the two domains, in contrast to the UNet model proposed in~\cite{Iwashita_2019_CVPR_Workshops}. However, the original CycleGAN is deterministic and incapable of generating multiple thermal images for one RGB input. 
ThermalGAN \cite{Kniaz_2018_ECCV_Workshops}, on the other hand, performs RGB2Thr translation with multiple outputs. Unlike CycleGAN, though, it is trained using supervision from matched pairs. 
In our work, we use a multimodal unsupervised I2I translation (MUNIT) model~\cite{huang2018munit} to be able to map a single RGB image to diverse thermal images.\\

\subsection{Varying output styles}
MUNIT~\cite{huang2018munit} introduces variations in its synthetic outputs by assigning them different random ``styles". This is achieved through the inclusion of AdaIN~\cite{huang2017adain} layers in the decoder. 
AdaIN adjusts the mean and standard deviation of features in a neural network layer, aligning them with a certain target style's statistics. 
This is done globally, across all positions in a feature map. 
As a result, different image components cannot be styled separately, limiting the potential of the framework for ADAS applications. 
One solution, as proposed in~\cite{shen2019towards}, is to apply separate AdaINs on different objects in a scene by relying on object bounding boxes and using multiple encoders and decoders. However, bounding boxes often still contain too much background information for a good result. 
Moreover, the need for additional 
decoders
makes the pipeline harder to train. 
Here, we leverage segmentation maps to apply the right styles to the right pixels. In addition, we propose a Component-aware AdaIN (CoAdaIN) that can easily be plugged into existing image translation frameworks, without requiring additional 
decoders.\\

\vspace{-5mm}
\section{Methodology}
Our goal is to produce more diverse thermal images from one RGB image.
We adopt the MUNIT \cite{huang2018munit} approach to achieve unpaired multi-modal RGB2Thr translation.
While MUNIT learns a bidirectional translation, 
we focus only on RGB2Thr in this paper. The Thr2RGB direction is co-optimised to enable unpaired training but is treated merely as an auxiliary stream here.
Different from MUNIT, we propose a novel adaptive instance normalization (AdaIN) block called CoAdaIN.
The proposed 
CoAdaIN enables the model to vary the style of different image components individually.
We denote the RGB and thermal modalities as $a$ and $b$, and their images as $I_a$ and $I_b$ respectively.


\subsection{Component-aware Adaptive Instance Normalization}
The original AdaIN~\cite{huang2017adain} transfers one style onto the entire target image, disregarding the possibility that different image components may exhibit different styles.
We introduce CoAdaIN to tackle this issue.
Firstly, we assume one image is composed of $K$ components, i.e. $I = \{I_{1}, I_{2}, I_{3}, ..., I_{K}\}$.  
The CoAdaIN operation is defined as follows:
\begin{align}
    CoAdaIN(x,y, i)=\sigma{(y_{i})} \frac{x_{i} - \mu{(x_{i})}}{\sigma(x_{i})} + \mu(y_{i})
    \label{eq::inadain}
\end{align}

\begin{figure}[t]
	\begin{minipage}[t]{0.48\textwidth}  
		\centering  
		\includegraphics[width=\textwidth]{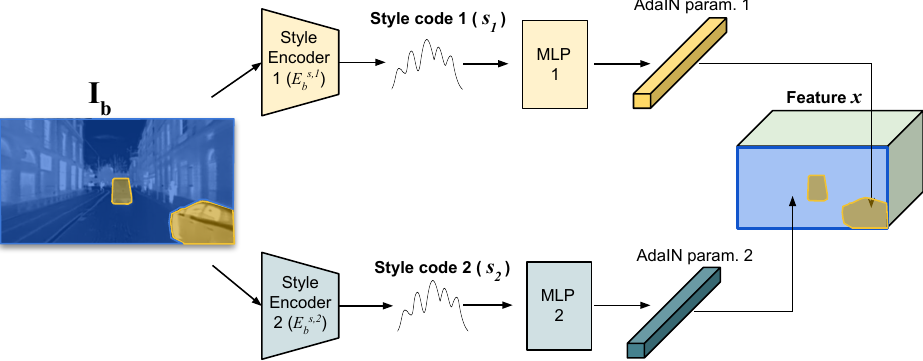}
	\end{minipage}  
	\centering  
    \vspace{-3mm}
	\caption{The illustration of CoAdaIN.
 For simplicity, here we consider two components: vehicles and background.
 During the training phase,
Style Encoder 1 (noted as $E^{s, 1}_b$) encodes the style code from the vehicle region (yellow mask). The style code is then sent to $MLP_1$ to generate CoAdaIN parameters (i.e. $\mu(y_{1})$ and $\sigma(y_{1})$).
Please note, $\mu(y_{1})$ and $\sigma(y_{1})$ are only applied to the corresponding region (i.e. yellow mask) in the feature $x$.
 A similar process applies to the background (blue mask).
 During the inference, the style codes ($s_1$ and $s_2$) are sampled from the Gaussian distribution.
 }
	\label{fig::inadain}
	\vspace{-5mm}
\end{figure}

$\mu(x_{i})$ and $\sigma(x_{i})$ as the mean and variance of the $i_{th}$ component in the feature tensor $x$. 
$\mu(y_{i})$ and $\sigma(y_{i})$ are computed via a multilayer perceptron (MLP) from each target style code $s_i$.
Fig. \ref{fig::inadain} shows an example of our CoAdaIN operation.
Original AdaIN 
overlooks the composite nature
of images, 
as it normalizes image features $x$ only per channel. 
By using CoAdaIN, we normalize $x$ not only per channel but also per component, enabling models to transfer different styles, such as thermal appearance, onto a single output image.

\begin{figure*}[htbp]
	\begin{minipage}[t]{1\textwidth}  
		\centering  
		\includegraphics[width=\textwidth]{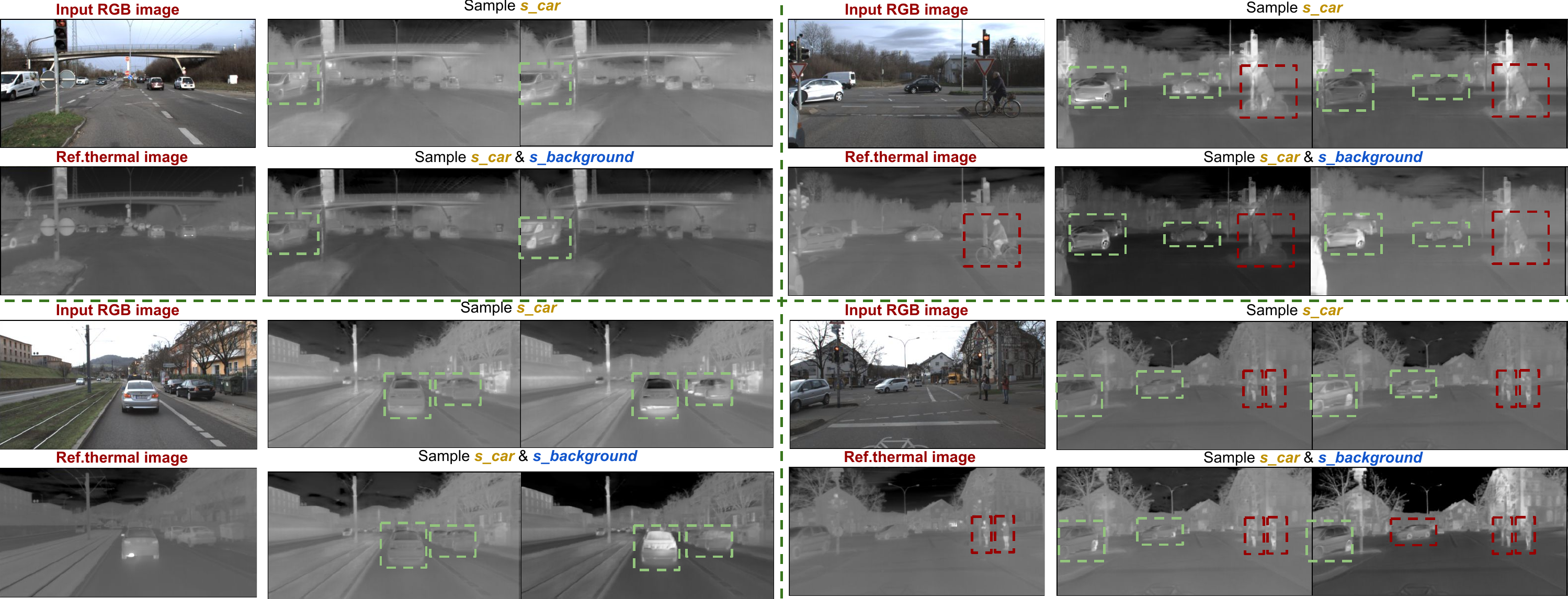}
	\end{minipage}  
	\centering  
 \vspace{-5mm}
	\caption{Qualitative results of our RGB2Thr translation. 
        For each group, the first row presents two translated images when only the style code for vehicles is sampled. 
    It is clear that the background thermal appearance remains the same while the vehicles' thermal appearance varies. 
    The second row shows the outputs when both vehicle and background style codes are sampled. In this case, the background's thermal appearance also changes, e.g. building, sky, road, etc. Please zoom in for more details.
        }
	\label{fig::qualitative}
	\vspace{-5mm}
\end{figure*}

\subsection{Model architecture and data flow}
Similar to \cite{huang2018munit}, we use two content encoders $E^{c}_{a}$, $E^{c}_{b}$ and two style encoder banks $E^{s}_{a}$, $E^{s}_{b}$ to encode the content and style codes of images from the RGB and thermal modalities.
Two generators, $G_{a}$ and $G_{b}$, decode the content and style codes to generate images in the RGB and thermal modalities, respectively. The semantic segmentation masks are defined as $M_a$ and $M_b$.
The full model consists of two streams, one for each translation direction. We focus the discussion on RGB2Thr as our main translation stream, but note that the auxiliary stream is symmetric to it.

$E^{c}_{a}$ encodes both RGB image $I_a$ and its semantic segmentation $M_a$ to obtain the content code $c_{a}$, namely $c_{a} = E^{c}_{a}(I_a\oplus M_a)$. 
Style codes $S_a=\{ s_1, s_2, \ldots, s_K \}$ are encoded via $E^{s, 1}_{a}$, $E^{s, 2}_{a}$, $...$,$E^{s, K}_{a}$ .
The reconstructed image $I_{aa}$ can then be obtained by decoding the extracted content and style code. $I_{aa} = G_{a}(c_{a}\oplus M_a, S)$.
Notation $\oplus$ refers to the concatenation operation.
For the cross-modality translation, the target style code $S_b$ is sampled from a Gaussian distribution, $S_b \sim N(0,1)$.
The translated image $I_{ab}$ is generated via $I_{ab} = G_{b}(c_a \oplus M_a, S_{b}).$
While segmentation maps were not included in \cite{huang2018munit}, they constitute an important input to the model in our work. They supplement the RGB inputs and help the model better understand the different image components.

\subsection{Learning process}
Besides the loss functions introduced in \cite{huang2018munit}, we include an improved diversity penalty loss \cite{ms_2019_Mao} to increase the diversity in the image components of interest, while keeping the overall images realistic.
\subsubsection{Object-centered diversity penalty loss}
\begin{align}
    \mathcal{L}_{\text{ocdp}} = (\frac{d_{I}(M_a*G_b(c_a, S_1), M_a*G_b(c_a, S_2)}{d_{S}(S_1,S_2)})^{-1}
    \label{eq::da_adv}
\end{align}
$M$ is the object mask,
and $S_1$, $S_2$ are two different style codes sampled from a Gaussian distribution.
We use $L 1$ distance for $d_I$ and $d_S$.
This loss encourages the model to generate a more diverse thermal appearance for the component of interest
when the content code is the same.

\section{Experiment}
\subsection{Dataset}
We use the Freiburg Thermal dataset\cite{vertens20bridging} to train and evaluate our model. The dataset contains 5-day driving sequences in the daytime, totalling 9,566 frames. The dataset consists of both RGB and thermal images as well as their semantic segmentation. The two-modality data have been synchronized and aligned.
We use the first 4 days' sequences to train our model (8,699 images) and the $5^{th}$ day's sequence (3,470 images) to test.
We follow \cite{vertens20bridging} to pre-process the images. The processed image size is $320 \times 700$ and we resize it to $256 \times 512$. 
For CoAdaIN, we define two components in our images, namely vehicles and background, therefore $K=2$.


\subsection{Qualitative result}
Fig. \ref{fig::qualitative} shows some of our RGB2Thr translation results.
It is clear that our method enables the model to generate different styles (i.e. thermal appearance) on specific components.
When we sample the style code for vehicles while keeping the background consistent, we can see that only the thermal appearance of vehicles varies. This is impossible in the original MUNIT model.
Please check the green dashed box for a more detailed thermal appearance of vehicles.

When also sampling the style code for the background, we can see different thermal appearances for buildings, sky, roads etc.
In addition, our model also preserves the pedestrian's thermal appearance well (see the red dashed box region on the top-right and bottom-right images).

\subsection{Quantitative Evaluation}
We trained the original MUNIT\cite{huang2018munit} model on this RGB2Thr task to serve
as our baseline.
We evaluate our model in two aspects: diversity and quality.
\subsubsection{Diversity}
 Learned Perceptual Image Patch Similarity (LPIPS)\cite{zhang2018perceptual} distance is widely used to measure the 
diversity in the outputs of a generative model.
Following \cite{ms_2019_Mao}\cite{DRIT}, we compute the average distance between 1,000 pairs of randomly translated images from 100 real images.
Higher LPIPS distance indicates better diversity among the generated images. 
Table \ref{table::quantitative} shows our model can reach higher LPIPS scores compared with the baseline.
We also compute the LPIPS distance for the case where we only sample the style code of vehicles while keeping the background style the same.
It cannot be computed for MUNIT since it fails to separate the style of different image components.

\subsubsection{Quality}
To evaluate the quality of our generated images, we adopt the Fréchet Inception Distance (FID) score\cite{Heusel2017GANsTB}.
We sample the style code three times for all the 3,470 testing images and compute the mean FID score with the standard deviation.
A lower FID value indicates better quality of the generated images. 
Table \ref{table::quantitative} shows
the FID score of our model is lower (30.91) than the baseline, 
indicating our generated images are
closer in distribution to the real images.


\begin{table}[!t]
\vspace{-5mm}
\renewcommand{\arraystretch}{1.3}
\caption{Quantitative results}
\vspace{-2mm}
\label{table:quantitative}
\centering
\begin{tabular}{|c||c|c|}
\hline
\bfseries Model & \bfseries MUNIT &  \bfseries Ours\\
\hline\hline
LPIPS $\uparrow$ & 0.1160 $\pm$ 0.0857 &  \textbf{0.1413} $\pm$ \textbf{0.0796}\\ 
LPIPS (vehicle) $\uparrow$ & - &  \textbf{0.0461} $\pm$ \textbf{0.0321}\\ 
\hline
FID $\downarrow$ & 33.86 $\pm$ 0.14 &   \textbf{30.91} $\pm$ \textbf{0.05}\\ 
\hline
\end{tabular}
\label{table::quantitative}
\vspace{-6mm}
\end{table}

\section{Conclusion and future work}
In this paper, we argue that RGB-to-thermal image translation should be thought of as a 
one-to-many mapping
problem
(i.e., one RGB input can correspond to multiple thermal outputs).
To improve the diversity of the output, we introduce the Component-aware Adaptive Instance normalization (CoAdaIN). 
This operation allows for different styles for different components.
In addition,
our quantitative result indicates that our method can efficiently increase the diversity and quality of the generated thermal images.
Next, we will focus on the synthetic RGB data generated by 3D simulators, such as CARLA\cite{carla}, which do not support thermal imaging simulation.
3D simulators can generate more diverse RGB images and create numerous of corner cases seldom encountered in real-world scenarios.
Therefore, it is worth exploring how to use our real-world image translation model to generate corresponding synthetic thermal images from synthetic RGB images.

\bibliographystyle{IEEEtran}
\bibliography{egbib}

\begin{thebibliography}{10}
\providecommand{\url}[1]{#1}
\csname url@samestyle\endcsname
\providecommand{\newblock}{\relax}
\providecommand{\bibinfo}[2]{#2}
\providecommand{\BIBentrySTDinterwordspacing}{\spaceskip=0pt\relax}
\providecommand{\BIBentryALTinterwordstretchfactor}{4}
\providecommand{\BIBentryALTinterwordspacing}{\spaceskip=\fontdimen2\font plus
\BIBentryALTinterwordstretchfactor\fontdimen3\font minus \fontdimen4\font\relax}
\providecommand{\BIBforeignlanguage}[2]{{%
\expandafter\ifx\csname l@#1\endcsname\relax
\typeout{** WARNING: IEEEtran.bst: No hyphenation pattern has been}%
\typeout{** loaded for the language `#1'. Using the pattern for}%
\typeout{** the default language instead.}%
\else
\language=\csname l@#1\endcsname
\fi
#2}}
\providecommand{\BIBdecl}{\relax}
\BIBdecl

\bibitem{wang2018pix2pixHD}
T.-C. Wang, M.-Y. Liu, J.-Y. Zhu, A.~Tao, J.~Kautz, and B.~Catanzaro, ``High-resolution image synthesis and semantic manipulation with conditional gans,'' in \emph{Proceedings of the IEEE Conference on Computer Vision and Pattern Recognition (CVPR)}, 2018.

\bibitem{CycleGAN2017}
J.-Y. Zhu, T.~Park, P.~Isola, and A.~A. Efros, ``Unpaired image-to-image translation using cycle-consistent adversarial networks,'' in \emph{Proceedings of the IEEE International Conference on Computer Vision (ICCV)}, 2017.

\bibitem{Kniaz_2018_ECCV_Workshops}
V.~V. Kniaz, V.~A. Knyaz, J.~Hladuvka, W.~G. Kropatsch, and V.~Mizginov, ``Thermalgan: Multimodal color-to-thermal image translation for person re-identification in multispectral dataset,'' in \emph{Proceedings of the European Conference on Computer Vision (ECCV) Workshops}, September 2018.

\bibitem{Liu_2018_CVPR_Workshops}
S.~Liu, V.~John, E.~Blasch, Z.~Liu, and Y.~Huang, ``Ir2vi: Enhanced night environmental perception by unsupervised thermal image translation,'' in \emph{Proceedings of the IEEE Conference on Computer Vision and Pattern Recognition (CVPR) Workshops}, June 2018.

\bibitem{Iwashita_2019_CVPR_Workshops}
Y.~Iwashita, K.~Nakashima, S.~Rafol, A.~Stoica, and R.~Kurazume, ``Mu-net: Deep learning-based thermal ir image estimation from rgb image,'' in \emph{Proceedings of the IEEE/CVF Conference on Computer Vision and Pattern Recognition (CVPR) Workshops}, June 2019.

\bibitem{Liu_HGGAN_2022}
T.~Liu, Y.~Liu, W.~Xu, Y.~Pu, Y.~Hao, and W.~Zuo, ``Hggan: Visible to thermal translation generative adversarial network guided by heatmap,'' in \emph{2022 IEEE International Conference on Unmanned Systems (ICUS)}, 2022, pp. 171--176.

\bibitem{huang2018munit}
X.~Huang, M.-Y. Liu, S.~Belongie, and J.~Kautz, ``Multimodal unsupervised image-to-image translation,'' in \emph{Proceedings of the European Conference on Computer Vision (ECCV)}, 2018.

\bibitem{huang2017adain}
X.~Huang and S.~Belongie, ``Arbitrary style transfer in real-time with adaptive instance normalization,'' in \emph{Proceedings of the IEEE International Conference on Computer Vision (ICCV)}, 2017.

\bibitem{escyclegan_2019_sun}
T.~Sun, C.~Jung, Q.~Fu, and Q.~Han, ``Nir to rgb domain translation using asymmetric cycle generative adversarial networks,'' \emph{IEEE Access}, vol.~7, pp. 112\,459--112\,469, 2019.

\bibitem{shen2019towards}
Z.~Shen, M.~Huang, J.~Shi, X.~Xue, and T.~Huang, ``Towards instance-level image-to-image translation,'' in \emph{Proceedings of the IEEE Conference on Computer Vision and Pattern Recognition (CVPR)}, 2019.

\bibitem{ms_2019_Mao}
Q.~Mao, H.-Y. Lee, H.-Y. Tseng, S.~Ma, and M.-H. Yang, ``Mode seeking generative adversarial networks for diverse image synthesis,'' in \emph{Proceedings of the IEEE Conference on Computer Vision and Pattern Recognition (CVPR)}, 2019, pp. 1429--1437.

\bibitem{vertens20bridging}
J.~Vertens, J.~Z{\"u}rn, and W.~Burgard, ``Heatnet: Bridging the day-night domain gap in semantic segmentation with thermal images,'' \emph{arXiv preprint arXiv:2003.04645}, 2020.

\bibitem{zhang2018perceptual}
R.~Zhang, P.~Isola, A.~A. Efros, E.~Shechtman, and O.~Wang, ``The unreasonable effectiveness of deep features as a perceptual metric,'' in \emph{Proceedings of the IEEE Conference on Computer Vision and Pattern Recognition (CVPR)}, 2018.

\bibitem{DRIT}
H.-Y. Lee, H.-Y. Tseng, J.-B. Huang, M.~K. Singh, and M.-H. Yang, ``Diverse image-to-image translation via disentangled representations,'' in \emph{European Conference on Computer Vision}, 2018.

\bibitem{Heusel2017GANsTB}
M.~Heusel, H.~Ramsauer, T.~Unterthiner, B.~Nessler, and S.~Hochreiter, ``Gans trained by a two time-scale update rule converge to a local nash equilibrium,'' in \emph{Advances in Neural Information Processing Systems (NeurIPS)}, 2017.

\bibitem{carla}
A.~Dosovitskiy, G.~Ros, F.~Codevilla, A.~Lopez, and V.~Koltun, ``{CARLA}: {An} open urban driving simulator,'' in \emph{Proceedings of the 1st Annual Conference on Robot Learning}, 2017, pp. 1--16.

\end{thebibliography}

\newpage
\end{document}